\documentclass[10pt, conference, letterpaper]{IEEEtran}
\usepackage[ruled,linesnumbered,vlined]{algorithm2e}
\usepackage{verbatim}
\usepackage{amsfonts}
\usepackage{epsfig}
\usepackage{algorithmic}
\usepackage{multicol}
\usepackage{multirow}
\usepackage{amsmath}
\usepackage{amssymb}
\usepackage{subfig}
\usepackage{tikz}
\usepackage{pgf}
\usepackage{float}
\usepackage{graphicx}
\usepackage{comment}
\usepackage{hyperref}

\usepackage{url}      

\makeatletter
\def\url@leostyle{%
  \@ifundefined{selectfont}{\def\UrlFont{\sf}}{\def\UrlFont{\small\bf\ttfamily}}}
\makeatother
\begin{document}

\title{Imbalanced Malware Images Classification: a CNN based Approach}

\author{\IEEEauthorblockN{Songqing Yue} 
\IEEEauthorblockA{University of Wisconsin\\
yues@uwplatt.edu}
\and  
\IEEEauthorblockN{Tianyang Wang} 
\IEEEauthorblockA{Austin Peay State University\\
toseattle@siu.edu}
}

\providecommand{\keywords}[1]{\textbf{\textit{Keywords---}} #1}

\maketitle

\begin{abstract}

Deep convolutional neural networks (CNNs) can be applied to malware binary detection via image classification. The performance, however, is degraded due to the imbalance of malware families (classes). To mitigate this issue, we propose a simple yet effective weighted softmax loss which can be employed as the final layer of deep CNNs. The original softmax loss is weighted, and the weight value can be determined according to class size. A scaling parameter is also included in computing the weight. Proper selection of this parameter is studied and an empirical option is suggested. The weighted loss aims at alleviating the impact of data imbalance in an end-to-end learning fashion. To validate the efficacy, we deploy the proposed weighted loss in a pre-trained deep CNN model and fine-tune it to achieve promising results on malware images classification. Extensive experiments also demonstrate that the new loss function can well fit other typical CNNs, yielding an improved classification performance.

\end{abstract}


\section{Introduction}
\label{introduction}

Malware binary, usually with a file name extension of ``.exe" or ``.bin",  is a malicious program that could harm computer operating systems. Sometimes, it may have many variations with highly reused basic patterns. This implies that malware binaries could be categorized into multiple families (classes), and each variation inherits the characteristics of its own family. Therefore, it is important to effectively detect malware binary and recognize possible variations \cite{demme2013feasibility,kolbitsch2009effective}.


This is non-trivial but challenging. A malware binary file can be visualized to a digital gray image \cite{nataraj2011malware}. After visualization, the malware binary detection turns into a multi-class image classification problem, which has been well studied in deep learning. 
One can manually extract features from malware images and feed them into classifiers such as SVM (support vector machine) or KNN (k-nearest neighbors algorithm) to detect malware binaries via classification. To be more discriminative, one can utilize CNN to automatically extract features as Razavian et al. did in \cite{sharif2014cnn} and perform classification in an end-to-end fashion. However, most deep CNNs are trained by properly solicited balanced data \cite{krizhevsky2009learning,ILSVRC15}, while malware image datasets may be highly imbalanced: some malware has many variations while some other only has few variations. For instance, the dataset \cite{nataraj2011malware} used in our paper consists of 25 classes, and some class includes more than 2000 images while some other only has 80 images or so. As a result,
the reputed pre-trained CNN models \cite{krizhevsky2012imagenet,simonyan2014very,szegedy2015going,he2016deep} may still perform poorly in this scenario. Furthermore, pre-trained CNN models are originally designed for specific vision tasks, and they cannot be applied to malware binary detection directly. 

One may argue that data augmentation could be a possible approach to balance data, such as oversampling minority classes and/or down-sampling majority classes. It is, however, not suitable for our problem due to two reasons. First, down-sampling may result in a loss of many representative malware variations. Second, simply augmenting data cannot guarantee to generate images corresponding to real malware binaries. To address the challenges, we investigate how to train a CNN model with the imbalanced data on hand. 

Inspired by the work in \cite{wang2017elu, huang2016learning} which designed new loss functions for improving CNN training, we propose a weighted softmax loss for deep CNNs on malware images classification. Based on the error rate given by the softmax loss, we weight misclassifications by different values corresponding to class size. Intuitively, misclassifications in minority classes should be amplified, and that in majority classes need to be suppressed. Our weighted loss can achieve this goal and guide the model weights updating in a proper direction. We adopt a pre-trained VGG-19 model \cite{simonyan2014very}, and retrain it to achieve promising results on the malware image classification. Once the proposed loss has been demonstrated effective on the VGG model, it can be extended to other models, such as GoogleNet \cite{szegedy2015going} and ResNet \cite{he2016deep}. 

The contributions of this work can be summarized into two-fold. First, we propose a weighted softmax loss to help CNNs deal with imbalanced data. Second, we apply the proposed loss to address the challenges in malware image classification. 


\section{Related Work}
\label{relatedwork}

\begin{figure}[!tbp]  
  \centering
  \subfloat[Malware image from the \emph{Adialer.C} family]
  {\includegraphics[width=0.22\textwidth]{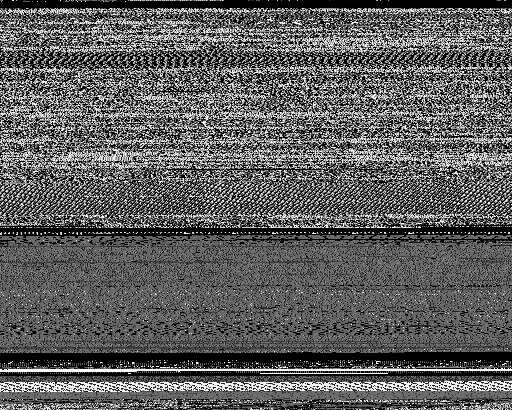}
  \label{fig1a}}
  \hfill
  \subfloat[Malware image from the \emph{Skintrim.N} family]
  {\includegraphics[width=0.22\textwidth]{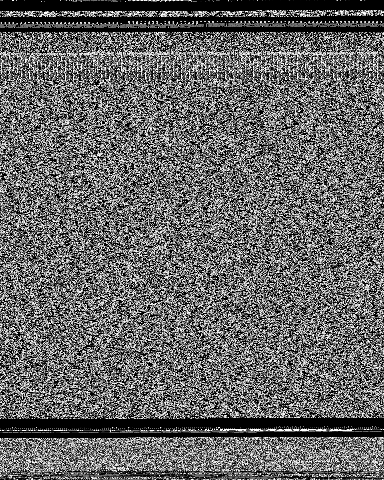}
  \label{fig:f2}}
  \caption{Two sample malware images.}
   \vspace{-0.1in}
  \label{MalwareImage}
\end{figure}

\textbf{Malware Images}: Nataraj et al. \cite{nataraj2011malware} proposed a way of converting malware binaries into digital gray images. They created 25 classes of malware images that are highly imbalanced. Two sample malware images from the \textquoteleft Adialer.C\textquoteright ~class and \textquoteleft Skintrim.N\textquoteright ~class are shown in Fig.~\ref{MalwareImage}. The Microsoft Malware Classification Challenge\footnote{https://www.kaggle.com/c/malware-classification} also provides imbalanced data which includes 9 classes only. Therefore, it is not as challenging as the one in \cite{nataraj2011malware} in terms of malware diversity. 

\textbf{Deep Learning}: Deep learning recently gains a remarkable success in computer vision. Krizhevsky et al. in \cite{krizhevsky2012imagenet} opened a new era for deep CNNs and their applications on image classification. Many works have been inspired thereafter. Simonyan and Zisserman proposed a very deep CNN in \cite{simonyan2014very}, and they significantly improved the performance by increasing the depth and using small convolutional filters with a size of $3\times3$. Szegedy et al. \cite{szegedy2015going} designed a 22 layers GoogLeNet which increased the width of the network, and achieved the new state-of-the-art performance on image classification. He et al. \cite{he2016deep} eased the training of a 152 layers deep CNN by presenting a residual learning scheme. 
Besides image classification, deep learning has also been applied on fundamental image processing tasks, such as denoising \cite{wang2017dilated} and contour detection \cite{bertasius2015deepedge}.  
Saxe and Berlin \cite{saxe2015deep} discussed the feasibility of using deep CNNs on malware detection. Hand-crafted features such as \textquoteleft PE import\textquoteright ~are extracted prior to the model training, which is not the typical manner of an end-to-end learning process. 
Huang et al. \cite{huang2016learning} proposed a quintuplet based triple header hinge loss for extracting discriminative features from imbalanced data. Nevertheless, their work requires extracting features in advance in a hand-crafted manner or using another pre-trained CNN model. Meanwhile, clustering algorithms such as k-means are needed prior to the training. These steps have not been integrated into an end-to-end fashion, employed by typical CNNs. 
Recently, Xu et al. \cite{xu2017deep} have applied deep learning in medical CT physics, and perform scatter correction via designing a new loss function.

\section{Weighted Softmax Loss}
\label{softmaxloss}

In this section, we introduce how to weight softmax loss according to class size. 

\subsection{Softmax Loss}

Softmax loss is a combination of softmax regression and entropy loss, used in multi-class classification problems. Given a $K$-classes training set including m images: \{$x^{(i)}$, $y^{(i)}$\}, where $x^{(i)}$ refers to an image, and $y^{(i)}$ denotes the ground truth label ($y^{(i)} \in \{1,2,…,K\}$). Let $a_j^{(i)}$ ($j=1,2,…,K$) be the output unit from the last fully connected layer of a CNN, then the probability that the label of $x^{(i)}$ is $j$ can be given by

\begin{equation}
    P_j^{(i)}=\frac{exp({a_j^{{i}}})}{\sum_{l=0}^K exp(a_l^{(i)})}.
\end{equation} Typical deep CNNs aim to minimize the entropy loss function: 

\begin{equation}
    J_0=-\frac{1}{m}\{ \sum_{i=1}^m \sum_{j=0}^K 1(y(i)=j) log~p_j^{(i)}\},
\end{equation}
where $m$ is the batchsize, and $K$ is the total number of the classes in a dataset. 1(.) is an indicator function such that, 1(true) gives 1, and 1(false) gives 0. In typical CNNs, convolutional filters are updated by the stochastic gradient descent (SGD) algorithm. The traditional softmax loss equally treats the misclassifications in each class, which is reasonable for balanced data, but will lead to a poor performance on imbalanced data, such as malware images. Our weighted softmax loss is proposed to address this issue. 

\subsection{Weighted Softmax Loss}
\label{PenalizedSoftmaxLoss}

Our approach can be formulated as follows, 

\begin{equation}
    J_0=-\frac{1}{m}\{ \sum_{i=1}^m \sum_{j=0}^K \omega_k *1(y(i)=j) log~p_j^{(i)}\},
\end{equation}
where $\omega_k$ is a weight coefficient determined by 

\begin{equation}
 \omega_k=1+\frac{S_{max}-S_k}{\beta*S_{max}},
\end{equation}  
where $k$ is the ground truth label of the $i^{th}$ image. $S_{max}$ is the size of the largest class in the dataset, and $S_k$ is the size of an arbitrary class $k$ in the dataset, and $\beta$ is a parameter that controls the scaling of the weighted loss. Our empirical preference of $\beta$ is 20. Extensive experiments about this parameter will be shown in section \ref{experiment}. It is worth noting that the minority classes will be assigned a larger weight coefficient whereas the majority classes will be lightly weighted. The weight coefficient will not dramatically affect the loss, thus it can be regarded as a subtle fine-tuning, that boosts the classification performance as well as avoids overfitting. In practice, the training procedure only needs to choose a $\omega_k$ according to the ground truth label of the $i^{th}$ sample in a mini-batch. 

\section{Classification with CNN}
\label{classification}

In this section, we discuss how to fine-tune the VGG-19 model for malware image classification.   

\subsection{Network Architecture}

Simonyan and Zisserman in \cite{simonyan2014very} proved that classification accuracy can be improved via deepening the network. The VGG-F model \cite{chatfield2014return} consists of 21 layers, whereas VGG-19 includes 43 layers. Convolutional filters with a size of $3\times3$ are used in VGG-19, and such smaller filters have been demonstrated being capable of extracting discriminative features. We fine-tune VGG-19 since it has been shown very competitive in the VGG family. 
Simonyan and Zisserman \cite{simonyan2014very} argued that the local response normalization (LRN) \cite{vedaldi2015matconvnet} could be ignored in CNNs. In order to boost the classification performance, we add Batch Normalization (BN) layer \cite{ioffe2015batch} between convolutional layer and ReLU (rectified linear unit) activation. However, the very first convolutional layer is followed by a ReLU layer directly, and the fully connected layer (expect the last one) is directly followed by a ReLU layer. 

A big potential threat to deep learning is overfitting, especially when a small training set is used. Srivastava et al. \cite{srivastava2014dropout} invented \emph{dropout}, a simple yet powerful approach to avoid overfitting in CNNs. Units in layers are randomly dropped and the corresponding connections are also removed temporarily. It works as randomly training different networks and averaging the results as the final one, aiming to enhance the generalization. We place two dropout layers (with probability $=0.5$) between the three fully connected layers to prevent overfitting (by default, the pre-trained VGG-19 does not have dropout layers). In testing stage, the dropout layers will be removed. 

\subsection{Weighted Softmax Loss Layer}

\begin{figure}[!htbp]
  \centering
 \includegraphics[width=0.5\textwidth]{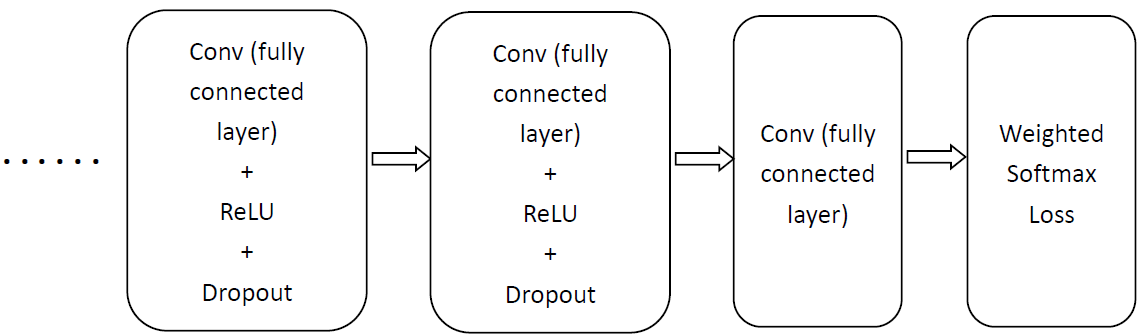}
  \vspace{-0.1in}
  \caption{Layer layout of our fine-tuned CNN.}
   \vspace{-0.1in}
  \label{figure2}
\end{figure}

We append the proposed loss as the last layer of our model. The final structure consists of 60 layers including the added dropout layers and BN layers. For brevity, only the fully connected layers, the added dropout layers, and the weighted loss layer are illustrated in Fig.~\ref{figure2}. 

\section{Experiments}
\label{experiment}

In this section, we validate the proposed loss on malware image classification. The VGG-F,M,S models \cite{chatfield2014return} are also employed to validate the general fit of the new loss function. We also analyze the selection of the scaling parameter and suggest an empirical option. The Top-1 validation error is adopted as the evaluation metric. 

We conduct the experiments with the MatConvNet\footnote{http://www.vlfeat.org/matconvnet/} framework \cite{vedaldi2015matconvnet}, which is an open source library for deep learning in matlab. One Nvidia TITAN X GPU is used to accelerate the mini-batch processing. 

\subsection{Dataset and Experimental Settings}

The dataset \cite{nataraj2011malware} used in our work consists of 25 classes which are highly imbalanced. The name and the size of each class is listed in Table \ref{25imbalance}\footnote{http://old.vision.ece.ucsb.edu/spam/malimg.shtml}. We partition the data as follows: the first 60\% images in each class are used for training, and the following 20\% for validation, and the last 20\% for testing. No data augmentation is applied, but we pre-process an input by subtracting its mean.    

\begin{table}[!htbp]
\centering
\caption{Names and sizes of the 25 imbalanced classes of malware images (Img stands for Images).}
\label{25imbalance}
\begin{tabular}{|c|c|c|c|}
\hline
No. & Type  & Family Name & \# of Img \\ \hline
1   & Worm  &  Allaple.L  &     1591        \\ \hline
2   & Worm  &  Allaple.A  &     2949        \\ \hline
3   & Worm  &  Yuner.A    &     800         \\ \hline
4   & PWS   & lolyda.AA 1 &     213         \\ \hline
5   & PWS   & lolyda.AA 2 &     184         \\ \hline
6   & PWS   & lolyda.AA 3 &     123         \\ \hline
7   & Trojan&  C2Lop.P    &     146               \\ \hline
8   & Trojan      &   C2Lop.gen!G          &   200              \\ \hline
9   & Dialer     &  Instantaccess           &       431          \\ \hline
10  & Trojan Downloader      &   Swizzor.gen!I          &     132            \\ \hline
11  & Trojan Downloader      &   Swizzor.gen!E         &      128           \\ \hline
12  & Worm      &   VB.AT          &    408             \\ \hline
13  & Rogue      &  Fakerean          &   381              \\ \hline
14  & Trojan      &   Alueron.gen!J          &   198              \\ \hline
15  & Trojan      &   Malex.gen!J          &     136            \\ \hline
16  & PWS      &   Lolyda.AT          &     159            \\ \hline
17  &  Dialer     &   Adialer.C          &     125            \\ \hline
18  &  Trojan Downloader     &     Wintrim.BX        &      97           \\ \hline
19  &  Dialer     &    Dialplatform.B         &     177            \\ \hline
20  &  Trojan Downloader     &  Dontovo.A           &      162           \\ \hline
21  &  Trojan Downloader    &  Obfuscator.AD           &     142            \\ \hline
22  &   Backdoor    &      Agent.FYI       &        116         \\ \hline
23  &  Worm:AutoIT     &     Autorun.K        &      106           \\ \hline
24  &  Backdoor     &    Rbot!gen         &      158           \\ \hline
25  &   Trojan    &    Skintrim.N         &      80           \\ \hline
\end{tabular}
\end{table}


\subsection{Effects of the Weighted Loss}
\begin{figure*}[!ht]
  \centering
  \subfloat[VGG-F model]
  {\includegraphics[width=0.3\textwidth]{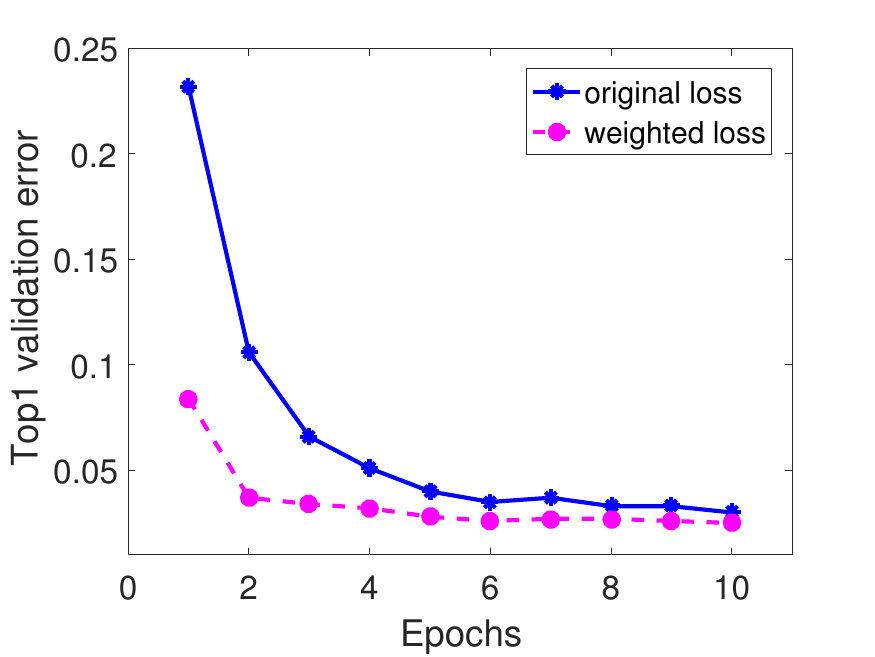}
  \label{fig3a}}
  \hfill
  \subfloat[VGG-M model]
  {\includegraphics[width=0.3\textwidth]{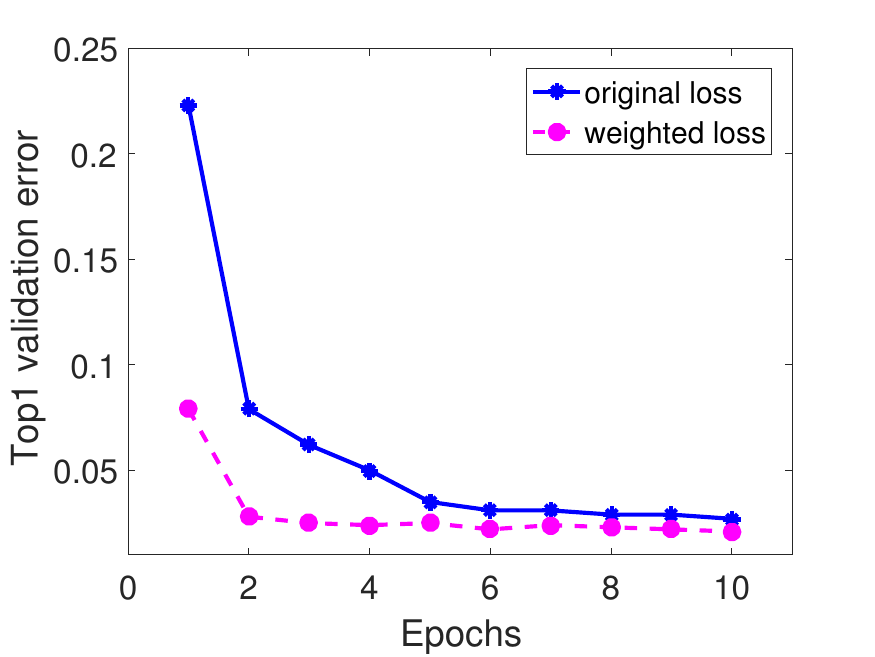}
  \label{fig3b}}
  \hfill
  \subfloat[VGG-S model]
  {\includegraphics[width=0.3\textwidth]{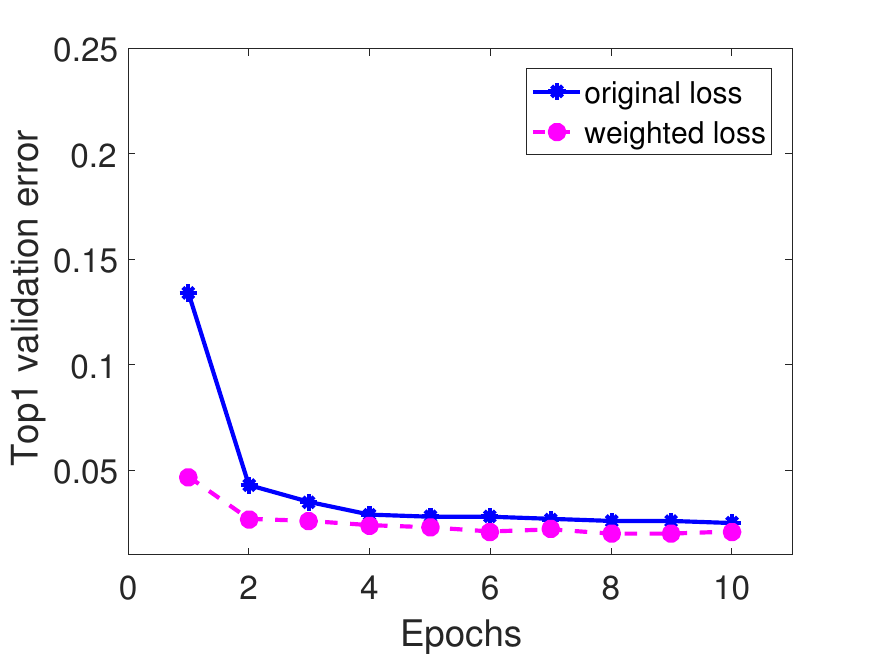}
  \label{fig3c}}
  \caption{Top-1 validation error of the VGG-F,M,S models with and without the weighted softmax loss.}
  \vspace{-0.1in}
  \label{figure3}
\end{figure*}

To validate the general fit, we deploy the proposed loss in 3 pre-trained CNN models, namely VGG-F, VGG-M, and VGG-S, respectively. The default structures are entirely preserved but the size of the last fully connected layer is changed from 1000 to 25, corresponding to the 25 classes in the malware dataset. We retrain the three networks, and the top-1 validation errors with and without the weighted loss are illustrated in Fig.~\ref{figure3}. It can be seen that our method effectively decreases the top-1 validation error, and keeps the curves stable. The test results are given in Table \ref{tableii}. 

\begin{figure*}[!ht]
  \centering
  \subfloat[Allaple.L]
  {\includegraphics[width=0.19\textwidth]{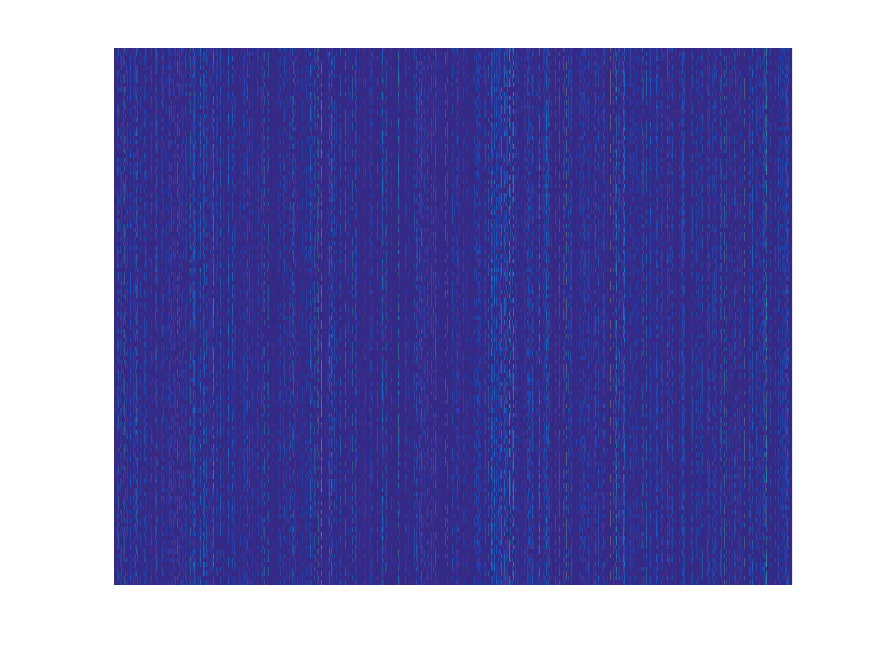}
  \label{fig5ab}}
  \hfill
  \subfloat[Adialer.C]
  {\includegraphics[width=0.19\textwidth]{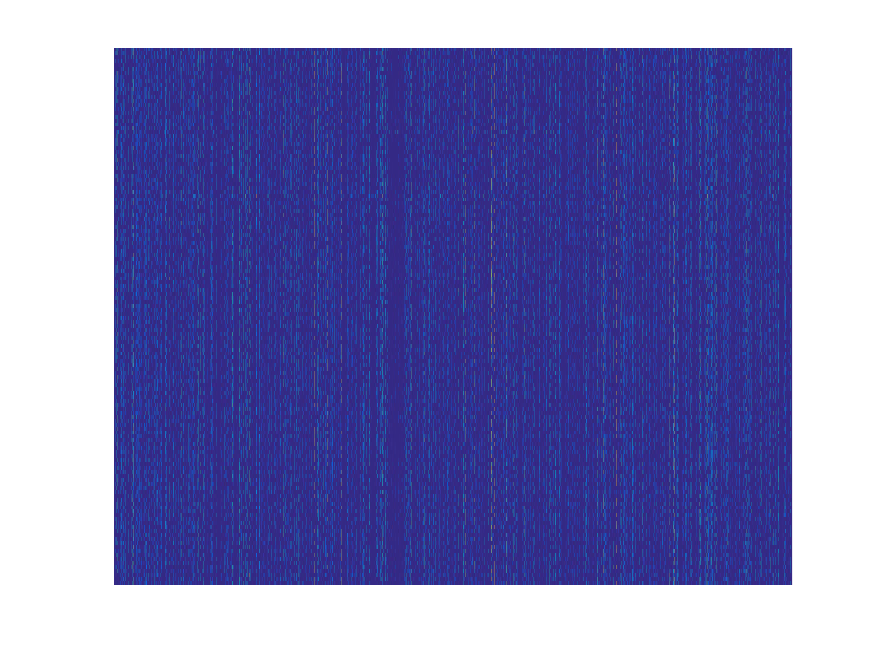}
  \label{fig5bb}}
    \subfloat[Yuner.A]
  {\includegraphics[width=0.19\textwidth]{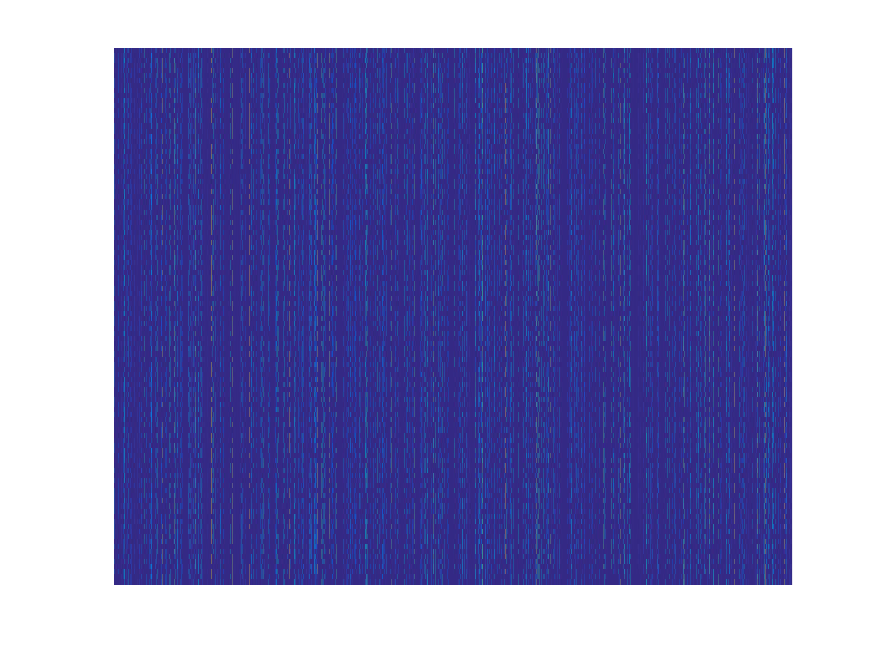}
  \label{fig5bc}}
    \subfloat[lolyda.AA 3]
  {\includegraphics[width=0.19\textwidth]{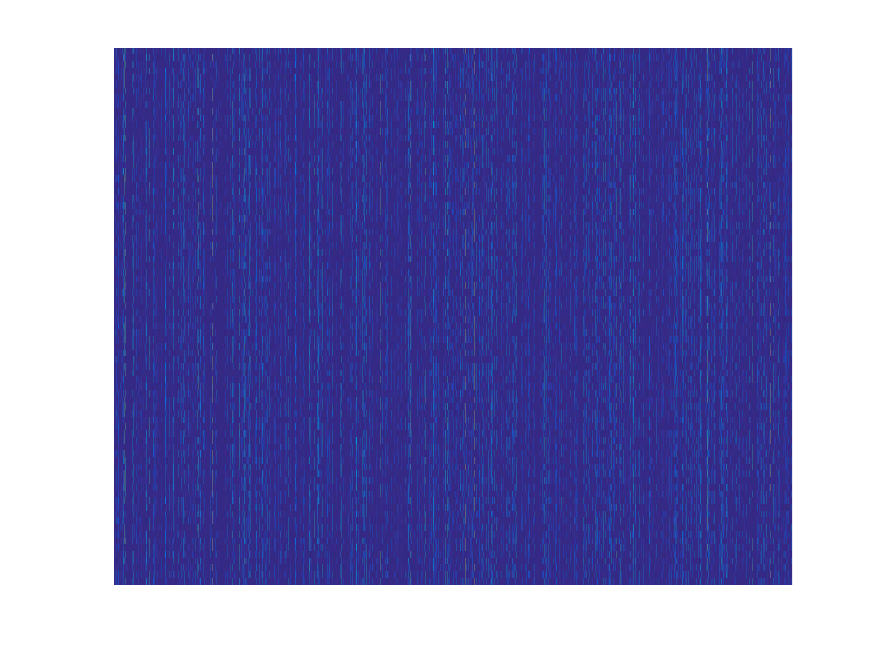}
  \label{fig5bd}}
    \subfloat[Autorun.K]
  {\includegraphics[width=0.19\textwidth]{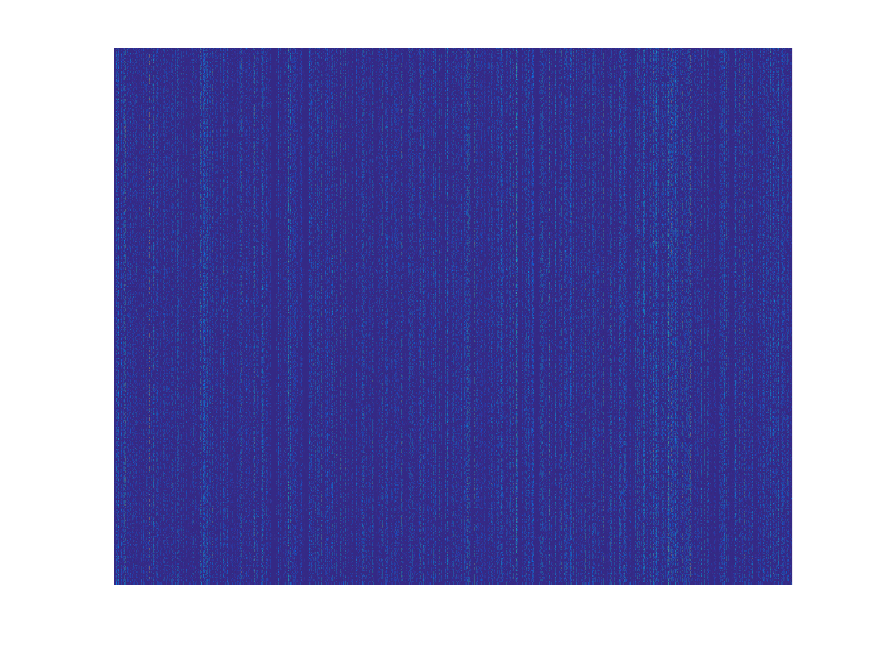}
  \label{fig5be}}
  
 
  \caption{Feature maps of five typical classes in the malware dataset.}
   \vspace{-0.1in}
  \label{figure5}
\end{figure*}

\begin{table}[]
\centering
\caption{Classification accuracy on the test images.}
\label{tableii}
\begin{tabular}{l|cc}

                       & Original loss & Weighted loss \\ \hline
VGG-19            & 97.32\%       & 98.63\%       \\ 
VGG-F                  & 94.48\%       & 95.36\%       \\ 
VGG-M                  & 95.90\%       & 96.39\%       \\ 
VGG-S                  & 96.23\%       & 96.89\%       \\ 
\end{tabular}
\end{table}

\subsection{Fine-tuning VGG-19 with the Weighted Loss}

\begin{figure}[!h]  
  \centering
 \includegraphics[width=0.4\textwidth]{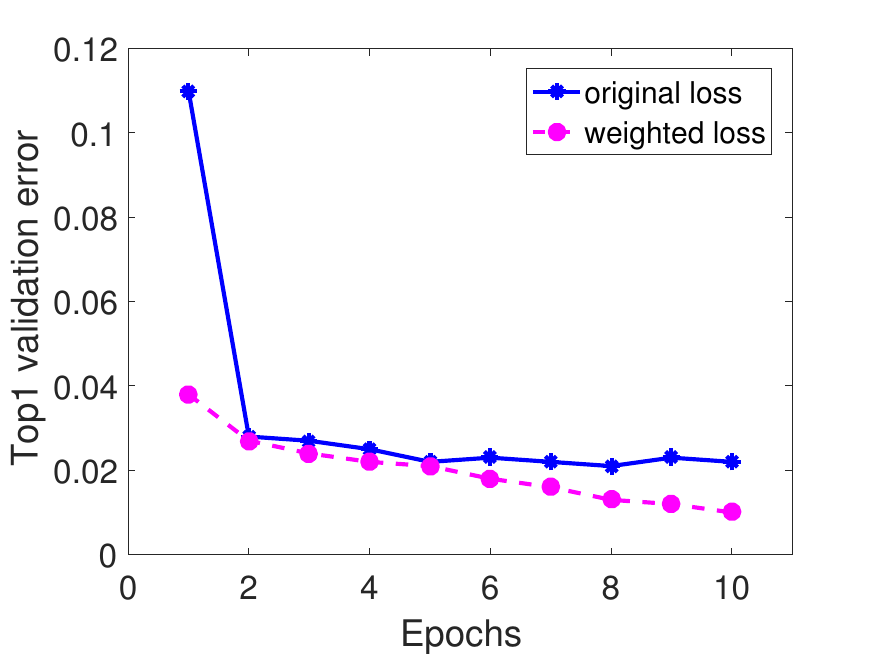}
  \caption{Top-1 validation error of our fine-tuned VGG-19 with and without the weighted softmax loss.}
   \vspace{-0.1in}
  \label{figure4}
\end{figure}

To achieve a better performance, we fine-tune VGG-19 as discussed in section \ref{classification}. We initialize the filter weights with MSRA \cite{he2016deep}. The momentum is set to 0.9, and the training is regularized by a weight decay rate of 0.0005. The learning rate is set to 0.0001, which we find leads to the best performance. Dynamic adjusting is not considered since our data size does not require many epochs to make the training converge. Batch size is set to 80. We compare the top-1 validation error of the models with and without the weighted loss. As shown in Fig.~\ref{figure4}, the weighted loss more effectively decreases the error compared to the original loss. During testing, we remove the two added dropout layers, and the results are presented in Table 2. In the dataset, the \textquoteleft Autorun.K\textquoteright, \textquoteleft Malex.gen!J\textquoteright, \textquoteleft Rbot!gen\textquoteright, \textquoteleft VB.AT\textquoteright, and \textquoteleft Yuner.A\textquoteright ~classes are from the same pack (UPX-Ultimate Packer for eXecutables). However, we treat them as different malware families. Otherwise, the test error can be further decreased. 

Moreover, CNNs can extract image features automatically. In our fine-tuned model, a 4096-dimensional feature vector for an arbitrary image can be captured at layer 41 (the second fully connected layer). We generate a feature map for each malware class by extracting the features for all the images in that class and combining all the feature vectors to form a matrix, which is finally visualized. Therefore, the dimension of any feature map is $n\times4096$, where $n$ is the class size. Such a feature map can reflect the characteristics of the corresponding malware class. We give the feature maps of five typical malware classes in Fig.~\ref{figure5}. Since the classes are highly imbalanced, $n$ is different for each class. To attain the best display effect, we use the `imagesc` command in matlab to visualize a feature map, automatically scaling the map.

  
 

\begin{figure}[!ht]
  \centering
 \includegraphics[width=0.4\textwidth]{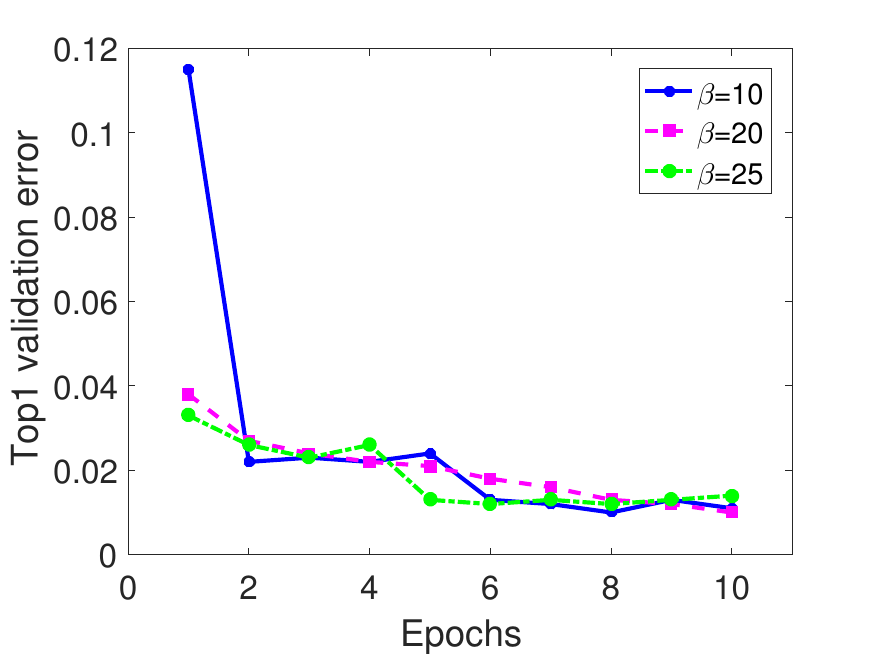}
  \caption{Impact of the scaling parameter.}
   \vspace{-0.1in}
  \label{figure6}
\end{figure}

\subsection{Scaling Parameter}


In Eq. (4), the parameter $\beta$ is needed to compute the weighted loss value. Empirically, 20 is the best option for $\beta$. To show the influence of this parameter, we further investigate three $\beta$ values on the fine-tuned VGG-19 and report the top-1 validation error in Fig.~\ref{figure6}. As can be seen, the curve with $\beta=20$ is smoother than the others and it also converges at a lower error rate. It indicates that an appropriate selection of $\beta$ is also critical to the network training.

\section{Conclusion}
\label{conclusion}

We propose a weighted softmax loss for convolutional neural networks on imbalanced malware image classification. By imposing a weight, the classification error for different classes can be treated unequally. The principle of weighting the loss has a clear intuition, and our experiments demonstrate its capacity of working with existing CNN models. We also fine-tune the pre-trained VGG-19 with the proposed loss to achieve a satisfactory classification performance. In addition, we have empirically suggested an option of the scaling parameter $\beta$ in computing the weighted loss. It indicates that an appropriate selection of this parameter is indispensable for the training success. \\ \\

  \bibliographystyle{unsrt}
  \bibliography{Reference} 

\end{document}